\documentclass[11pt,a4paper]{article}
\usepackage{arxiv_layout}
\usepackage{amssymb,booktabs,graphicx,tikz,array}

\usepackage{amsmath,amsfonts,bm}

\def\eqref#1{equation~\ref{#1}}

\def\1{\bm{1}}

\DeclareMathAlphabet{\mathsfit}{\encodingdefault}{\sfdefault}{m}{sl}
\SetMathAlphabet{\mathsfit}{bold}{\encodingdefault}{\sfdefault}{bx}{n}

\usepackage[authoryear,round]{natbib}
\usepackage{url,enumitem,placeins,float}
\usepackage[breaklinks,colorlinks,linkcolor=linkblue,citecolor=linkblue,urlcolor=linkblue]{hyperref}
\hypersetup{
  pdftitle={LabWorld: Learning Interactive 4D Worlds for Scientific Experimentation},
  pdfauthor={Suozhi Huang, Kelsey Fu, Binxu Li, Luna Lyu, Lezhi Tan, Kaize Wu, Wenbo Lu, David Smerkous, Jiahan Li, Siqi Zhu, Zhizhou Sha, Jiacheng Guo, Zhixuan Liang, Shilong Liu, Le Cong, Mengdi Wang}
}
\usetikzlibrary{arrows.meta,positioning}

\newcommand{\method}{\textsc{LabWorld}}
\newcommand{\SEthree}{\mathrm{SE}(3)}
\definecolor{capabilitygreen}{RGB}{25,125,65}
\definecolor{capabilitygray}{RGB}{115,115,115}
\newcommand{\capyes}{\textcolor{capabilitygreen}{\ensuremath{\checkmark}}}
\newcommand{\capno}{\textcolor{capabilitygray}{\ensuremath{\times}}}

\newcommand{\myparagraph}[1]{\vspace{2pt}\textbf{#1}}

\title{\textsc{LabWorld}: Learning Interactive 4D Worlds for Scientific Experimentation}

\author{%
\begin{minipage}{\textwidth}
\centering
{\fontsize{10}{13}\selectfont\bfseries
Suozhi Huang\textsuperscript{1,*}, Kelsey Fu\textsuperscript{1,*}, Binxu Li\textsuperscript{1}, Luna Lyu\textsuperscript{2}, Lezhi Tan\textsuperscript{2}, Kaize Wu\textsuperscript{3},\\
Wenbo Lu\textsuperscript{4}, David Smerkous\textsuperscript{5}, Jiahan Li\textsuperscript{4}, Siqi Zhu\textsuperscript{6}, Zhizhou Sha\textsuperscript{7},\\
Jiacheng Guo\textsuperscript{1}, Zhixuan Liang\textsuperscript{1}, Shilong Liu\textsuperscript{1}, Le Cong\textsuperscript{2}, Mengdi Wang\textsuperscript{1}\endgraf}
\vspace{6pt}
{\fontsize{9}{12}\selectfont
\textsuperscript{1}Princeton University \quad
\textsuperscript{2}Stanford University \quad
\textsuperscript{3}University of Chicago\\
\textsuperscript{4}New York University \quad
\textsuperscript{5}Oregon State University\\
\textsuperscript{6}University of Illinois at Urbana-Champaign \quad
\textsuperscript{7}University of Texas at Austin\endgraf}
\vspace{4pt}
{\fontsize{9}{11}\selectfont
\textsuperscript{*}These authors contributed equally to this work.\endgraf}
\end{minipage}%
}
\date{}

\begin{document}
\maketitle

\begin{abstract}
Autonomous scientific agents must not only reason about experiments but
also interact with the physical environments in which they are executed.
Existing laboratory simulators typically rely on manually constructed
assets and predefined procedures, while laboratory videos capture real
equipment, layouts, and interactions but leave this information implicit
in pixels. We introduce \method, an agent-driven Real2Sim framework for reconstructing interactive 4D laboratory worlds from monocular RGB video. \method\
recovers reusable 3D assets, a metric scene, time-varying object states,
and procedure events. Multi-frame observations provide complementary
evidence for asset and scene reconstruction, while motion reconstruction
combines visual and geometric observations with event constraints encoding
the spatial and temporal requirements of laboratory interactions. We evaluate \method\ on real pipetting and vortexing videos collected in our laboratory against
general-purpose agentic and dedicated reconstruction baselines.
For pipetting, \method\ achieves a mean Mask IoU of \textbf{0.501},
exceeding the strongest baseline by 0.033.
For vortexing, it achieves a mean Tube-Length--Normalized PCK-AUC of
\textbf{0.638}, exceeding the strongest baseline by 0.035.
On five pipetting videos, event-conditioned reconstruction increases
the fraction of modeled physical constraints satisfied from
\textbf{14.3\%} (5/35) to \textbf{100\%} (35/35).
The resulting worlds expose geometry, motion, articulation,
and events as explicit, editable state for replay, inspection, and future
embodied scientific systems.

\end{abstract}

\section{Introduction}
\label{sec:introduction}

Autonomous scientific discovery requires AI agents to connect experimental reasoning to physical execution~\citep{fu2026agenticlaboratories}.
A written protocol specifies what to do, but leaves implicit the state of a particular laboratory for the robot execution, including where equipment is located, how it is configured, and how objects must move and interact during execution.
Therefore, an environment representation is required to make these conditions explicit. An interactive 4D world provides such a representation by combining an editable 3D scene with time-varying object states and interaction events, allowing an observed procedure to be inspected, modified, and replayed.

\begin{figure*}[t]
    \centering
    \includegraphics[width=\textwidth]{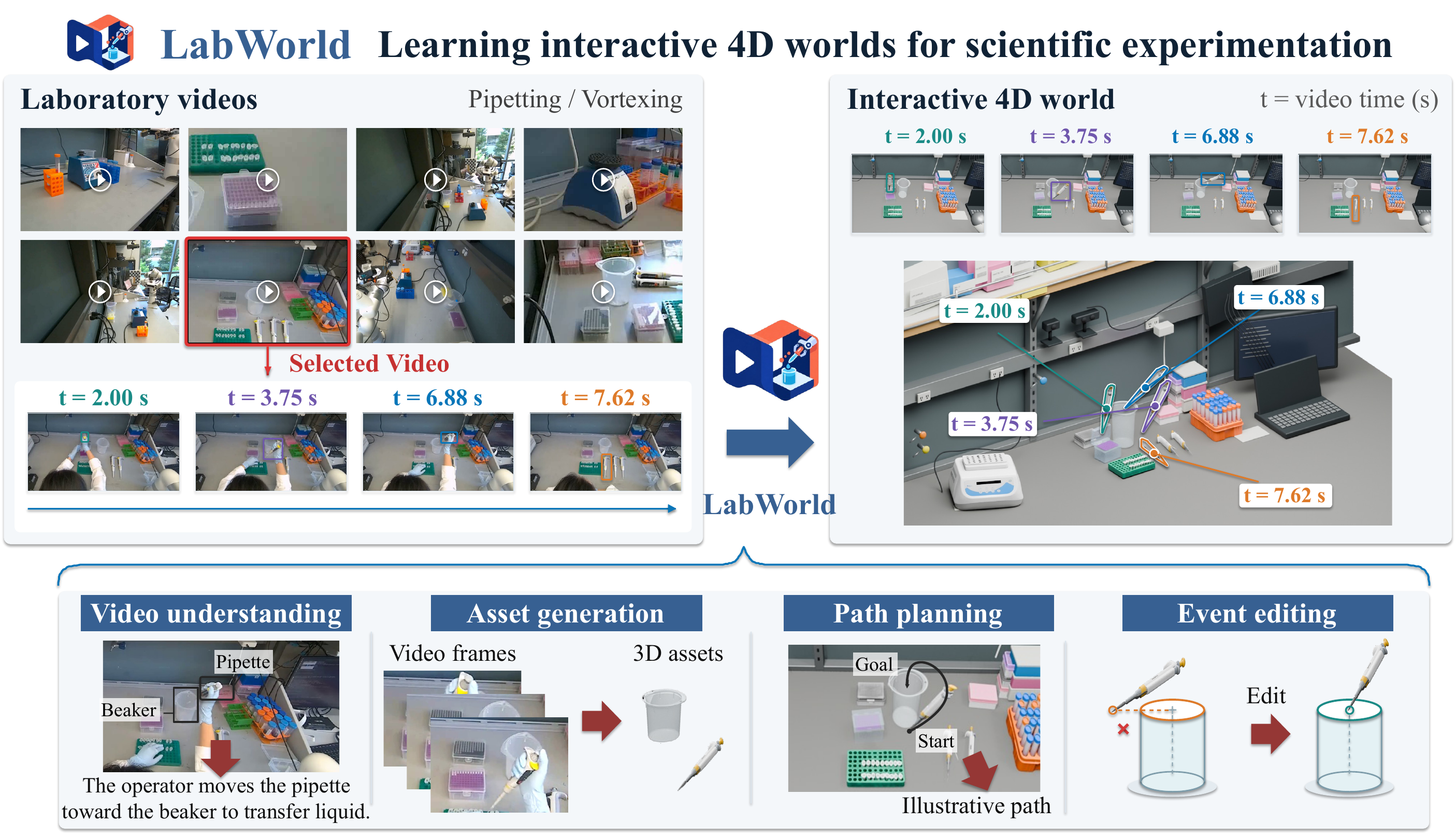}

    \caption{\textbf{From laboratory video to interactive 4D worlds.}
    Each video is selected from the collection on the left. Source frames
    are paired with reconstructed views at matching video times $t$, measured
    in seconds. The enlarged scene overlays the same pipette, with matching colors linking its reconstructed states to the source
    frames. The lower panels illustrate video understanding, reusable asset
    generation, and potential uses in path planning and event editing. Object
    boxes and interaction arrows in the video-understanding panel are
    explanatory annotations. The planning and editing panels are conceptual
    illustrations rather than evaluated downstream results.}
    \label{fig:system-overview}

\end{figure*}

Constructing such worlds currently requires substantial engineering.
Laboratory simulators rely on curated assets and predefined
procedures~\citep{lan2025autobio}, while many Real2Sim systems assume
existing asset libraries, depth sensing, or active
capture~\citep{torne2024rialto,pfaff2025scalablereal2sim}.
Human laboratory videos already contain much of the information that this
engineering supplies, like the equipment, its layout, and its use throughout
a procedure, but only implicitly in image sequences. Recent video-based
Real2Sim methods provide an important foundation
~\citep{ranawaka2026simfoundry,chen2026agenticreal2sim}, but do not jointly
target reusable laboratory assets, observed object motion, and
procedure-specific interaction constraints from ordinary monocular
laboratory video.

We introduce \method, an agent-driven framework that converts monocular
laboratory video into an explicit, editable 4D world (Fig.~\ref{fig:system-overview}). Laboratory objects
are often small, repetitive, and partially occluded, so \method\ combines
complementary observations across time to construct reusable 3D assets.
Metric geometry and camera estimation place these assets in a common scene,
while visual tracking and object geometry provide evidence for time-varying
rigid and articulated states. Unlike a generated video replay, the resulting
world exposes geometry, camera state, object trajectories, articulation,
and events as independently inspectable variables.

However, visual agreement alone does not establish that an experimental
interaction has been reconstructed correctly. A trajectory can align well in image space
while missing a vessel opening, intersecting surrounding geometry, or
violating support under monocular depth
ambiguity and hand occlusion. We propose \textit{event-conditioned reconstruction} to address this gap by constraining object motion with the spatial and temporal requirements of procedure events extracted from the video. For example, a pipette--beaker interaction requires
an approach, entry, dwell, and withdrawal through the opening rather than
merely placing the pipette near the beaker in projection. This distinction
between image agreement and interaction structure is central to our
reconstruction objective.

We collect an evaluation dataset of real pipetting and vortexing videos in our laboratory.
We evaluate \method{} on this dataset against general-purpose agentic and dedicated reconstruction baselines, including GPT-6 Astra, FoundationPose, and Do as I Do.
\method\ achieves the highest mean Mask IoU for pipetting
(\textbf{0.501}) and the highest mean Tube-Length--Normalized PCK-AUC
for vortexing (\textbf{0.638}) across all evaluated baselines. More importantly, our ablations expose a
limitation of image-space evaluation alone: on a five-video ablation
subset, removing event-conditioned
reconstruction increases pipette Mask IoU from 0.508 to 0.572 while
reducing satisfaction of the modeled physical requirements from
\textbf{35/35} to \textbf{5/35}. These diagnostics do not constitute
independent verification of hidden source events, but show that better
2D agreement can correspond to a less physically valid reconstruction.

Our main contributions are threefold.

\begin{itemize}[leftmargin=*, labelindent=0pt, itemsep=0pt]
    \item We present \method{}, an agent-driven Real2Sim framework that reconstructs an explicit 4D laboratory world from monocular video, comprising reusable assets, a metric scene, time-varying object states, and procedure events.

    \item We propose event-conditioned reconstruction that combines visual and geometric evidence with the spatial and temporal requirements of laboratory interactions, constraining solutions that are visually plausible but physically inconsistent.

    \item We collect a real laboratory video dataset covering pipetting and vortexing, and evaluate \method{} against agentic and dedicated reconstruction baselines.
    Component ablations examine the roles of visual refinement, geometric initialization, temporal regularization, and event-conditioned reconstruction.
\end{itemize}

\section{Related Work}
\label{sec:related-work}

\begin{table}[t]
    \centering
    \small
    \setlength{\tabcolsep}{3.2pt}
    \caption{
        Capabilities of related methods and evaluated baselines.
        Columns denote reusable 3D assets, reconstruction of the observed
        scene and motion, and explicit procedure-level constraints during
        motion reconstruction. \capyes\ indicates a demonstrated capability
        under these definitions; \capno\ indicates that it is not demonstrated.
        For GPT-6 Astra, the entries describe the video-based reconstruction agent evaluated in this study rather than the underlying model in isolation.
        Its inputs and reconstruction procedure are described in Appendix~\ref{app:baseline-details}.
    }
    \label{tab:related-work}
    \vspace{4pt}
    \resizebox{1\textwidth}{!}{
    \begin{tabular}{lcccc}
        \toprule
        Method
        & \shortstack{Laboratory\\assets}
        & \shortstack{Scene\\reconstruction}
        & \shortstack{Motion\\reconstruction}
        & \shortstack{Event-conditioned\\reconstruction} \\
        \midrule
        AutoBio~\citep{lan2025autobio}
        & \capyes & \capno & \capno & \capno \\
        SafeLab~\citep{bai2026safelab}
        & \capyes & \capno & \capno & \capno \\
        TableVerse~\citep{wang2026tableverse}
        & \capno & \capyes & \capno & \capno \\
        SimFoundry~\citep{ranawaka2026simfoundry}
        & \capno & \capyes & \capno & \capno \\
        Agentic Real2Sim~\citep{chen2026agenticreal2sim}
        & \capno & \capyes & \capyes & \capno \\
        \midrule
        GPT-6 Astra
        & \capyes & \capyes & \capyes & \capno \\
        FoundationPose~\citep{wen2024foundationpose}
        & \capno & \capno & \capyes & \capno \\
        Do as I Do~\citep{paliwal2026doasido}
        & \capyes & \capno & \capyes & \capno \\

        \midrule
        \method{} (ours)
        & \capyes & \capyes & \capyes & \capyes \\
        \bottomrule
    \end{tabular}}
\end{table}

We organize related work around laboratory simulation, Real2Sim
reconstruction, and the perception and representation components needed
to recover interactive 4D worlds.

\subsection{Laboratory Simulators and Digital Twins}

General manipulation benchmarks provide reusable tasks and physics
environments~\citep{yu2019metaworld,zhu2020robosuite,
gu2023maniskill2,mu2025robotwin}, while laboratory benchmarks extend
simulation to scientific equipment, precision manipulation, and
domain-specific state changes
~\citep{li2024chemistry3d,li2025labutopia,lan2025autobio,
bai2026safelab,darvish2026matterix,wu2026labimus,
cao2026labbuilder,liu2026pipette,tang2026labdex}.
These systems provide increasingly capable scientific environments, but
primarily construct them from curated assets, protocols, or engineered
workstations rather than recover the time-varying state of an observed
human procedure.

A-Lab and Coscientist demonstrate closed-loop scientific reasoning and
robotic execution~\citep{szymanski2023alab,boiko2023coscientist}, while
LabOS connects laboratory perception to human experimenters through XR
guidance~\citep{cong2025labos}. \method\ addresses a complementary
representation problem: converting recorded experiments into explicit
assets, scene state, motion, and interaction structure.

\subsection{Real2Sim and Motion Reconstruction}

Real2Sim systems reconstruct simulation state from real observations.
RialTo, Scalable Real2Sim, and Real-is-Sim recover workspaces or dynamic
twins for policy learning and evaluation
~\citep{torne2024rialto,pfaff2025scalablereal2sim,
abouchakra2025realissim}, but rely on active capture, RGB-D observations,
robot state, or controlled interaction beyond ordinary monocular video.
TableVerse, SimuScene, and RoboSnap reduce these requirements for
simulation-ready scene reconstruction
~\citep{wang2026tableverse,lee2026simuscene,zhang2026robosnap}.

Closer to our setting, SimFoundry converts short videos into physics-ready
scenes and Agentic Real2Sim uses vision-language agents to reconstruct
simulatable interaction episodes
~\citep{ranawaka2026simfoundry,chen2026agenticreal2sim}.
Human-video methods instead focus on manipulated-object motion:
Video2Sim2Real transfers object-centric keyframes to robot skills, and
Do as I Do reconstructs object geometry and 6-DoF trajectories from
monocular human video
~\citep{han2026video2sim2real,paliwal2026doasido}.
FoundationPose provides general 6D pose estimation and tracking from an
object model or reference views~\citep{wen2024foundationpose}.
\method\ differs by reconstructing manipulated-object motion within an
explicit laboratory world and using procedure events to constrain
ambiguous trajectories.

\subsection{Assets, Geometry, and Constraint Representations}

Interactive worlds require persistent objects and interaction structure,
not only appearance or correspondence. NeRF and 3D Gaussian Splatting
provide view synthesis~\citep{mildenhall2020nerf,kerbl2023gaussians};
DUSt3R, VGGT, VGGT-$\Omega$, Depth Anything 3, and MapAnything recover
camera and scene geometry
~\citep{wang2024dust3r,wang2025vggt,wang2026vggtomega,
lin2025depthanything3,keetha2025mapanything}; and CoTracker3,
SpatialTrackerV2, Track4World, and 4RC provide temporal correspondence
~\citep{karaev2024cotracker3,xiao2025spatialtrackerv2,
lu2026track4world,luo2026fourrc}. These signals provide geometric evidence
but do not alone define reusable objects, articulation, or functional
interaction requirements.

SAM2 and SAM3 provide prompted object segmentation
~\citep{ravi2024sam2,carion2025sam3}; SF3D, TRELLIS, and SAM3D recover
textured assets; and Particulate, Articraft, and RORA extend reconstruction
to articulated objects
~\citep{boss2024sf3d,xiang2025trellis,sam3dteam2025,
li2025particulate,zhou2026articraft,lee2026rora}.
ReKep represents manipulation objectives as relational keypoint
constraints~\citep{huang2024rekep}. \method\ integrates these capabilities
at the world level: reusable laboratory assets and metric geometry provide
persistent state, while source-video events constrain rigid and articulated
motion.

\section{Method}
\label{sec:method}

\myparagraph{Overview.}
Given a monocular RGB video of a laboratory procedure
$\mathcal{V}=\{I_t\}_{t=1}^{T}$, \method\ is designed to recover an
explicit laboratory world
\begin{equation}
    \mathcal{W}
    =
    \left(
        \mathcal{A},
        \mathcal{S},
        \mathcal{C}_{1:T},
        \mathcal{Q}_{1:T},
        \mathcal{E}
    \right),
    \label{eq:world}
\end{equation}
where $\mathcal{A}$ is a reusable asset library, $\mathcal{S}$ is the
metric static scene, $\mathcal{C}_{1:T}$ contains camera intrinsics and
extrinsics, $\mathcal{Q}_{1:T}$ contains time-varying rigid poses and
articulated states, and $\mathcal{E}$ is a structured event list.

\begin{figure*}[t]
    \centering
    \includegraphics[width=\textwidth]{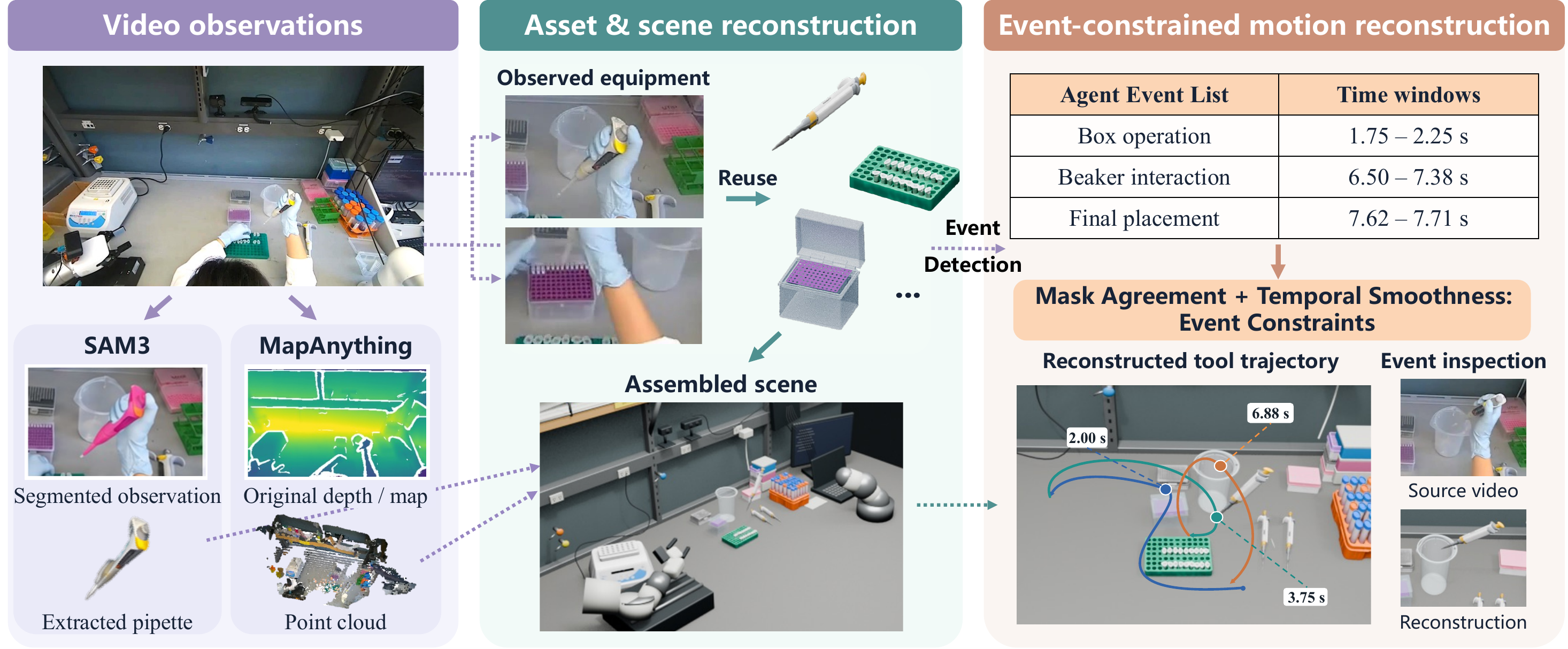}

    \caption{\textbf{Overview of the LabWorld reconstruction pipeline.}
    SAM3 provides visible object masks, while MapAnything estimates scene
    geometry from the source video. Observed equipment is represented by
    generated or reused assets and assembled into a shared 3D scene.
    An agent-generated event list specifies interaction types and time windows
    for motion reconstruction, which combines visual and geometric observations
    with temporal regularization and event constraints. The colored curve shows
    the reconstructed pipette-tip trajectory, with markers identifying selected
    times. Source-video and reconstruction crops support visual inspection of a beaker
    interaction, without independently establishing event completion.}
    \label{fig:method-details}
\end{figure*}

As illustrated in Fig.~\ref{fig:method-details}, the pipeline
(1) identifies objects and procedure events, (2) constructs reusable
assets and a metric scene, and (3) reconstructs motion under visual,
geometric, temporal, event, and physical constraints. A vision-language
agent prepares task-relevant object and event information from the source
video; segmentation, metric geometry, asset generation, and trajectory
optimization then recover the corresponding world state. The complete
workflow is designed for end-to-end automation. Appendix~\ref{app:automation-scope} separately documents the provenance and
scope of the cached-input reproduction runs used for reproducibility.

\subsection{Object Identification and Event List Generation}

\label{sec:video-understanding}

A vision-language model identifies task-relevant entities and functional
parts, while a promptable video segmenter propagates object identity
through the sequence to produce visible masks $M_t^i$. Multi-frame
association distinguishes repeated instances such as pipettes, tubes,
racks, and tip boxes.
The agent additionally infers a structured event list
\begin{equation}
    \mathcal{E}=\{e_k\}_{k=1}^{K},
    \qquad
    e_k=(\tau_k,\mathcal{P}_k,[t_k^s,t_k^e],r_k,c_k),
    \label{eq:event-graph}
\end{equation}
where $\tau_k$ is the event type, $\mathcal{P}_k$ its participating
entities, $[t_k^s,t_k^e]$ its temporal extent, $r_k$ a functional spatial
relation, and $c_k$ its confidence. Candidate events are proposed from
temporally sampled video and refined using object motion, mask containment,
relative distance, velocity changes, and articulation cues. The vocabulary
includes grasp, transport, place, insert, dispense, open, close, and
release. These events are not merely semantic labels: they instantiate
geometric conditions for reconstruction. For example,
\texttt{dispense(pipette, beaker)} associates the pipette tip with the
beaker opening and defines the interval over which the corresponding
interaction constraints apply. Additional event definitions are given in
Appendix~\ref{app:event-constraints}.

\subsection{Laboratory Assets and Scene Reconstruction}

\label{sec:asset-scene}

For each tracked object, \method\ selects complementary observations that
favor visibility, sharpness, viewpoint diversity, and consistency with
the tracked instance. A generative canonicalizer removes hands and scene
clutter, and a multi-view image-to-3D generator reconstructs a textured
asset. Each asset is assigned a canonical frame, physical scale, support
surface, collision geometry, and functional landmarks such as tips,
openings, slots, and hinges. Existing assets are reused when matched to
the laboratory library; articulated equipment is represented as links and
joints rather than a monolithic mesh.

MapAnything~\citep{keetha2025mapanything} estimates per-frame depth
$D_t$, point maps, and camera intrinsics and extrinsics
$(K_t,T^W_{C,t})$, lifting visual observations into a common world frame.
The dominant work surface defines the laboratory vertical and support
height, while known equipment dimensions establish metric scale. Static
asset poses are estimated jointly from multi-view reprojection, depth,
support, collision, and scale evidence, producing a metric scene shared
across observations.

\subsection{Motion Reconstruction with Event Constraints}

\label{sec:event-dynamics}

For each manipulated object, masks, functional landmarks, and metric
geometry initialize its time-varying state. Rather than estimating each
frame independently, we optimize the complete trajectory:
\begin{align}
    \mathcal{Q}_{1:T}^{*}
    =
    \arg\min_{\mathcal{Q}_{1:T}}\;&
    \lambda_{\mathrm{obs}}\mathcal{L}_{\mathrm{obs}}
    +\lambda_{\mathrm{geo}}\mathcal{L}_{\mathrm{geo}}
    +\lambda_{\mathrm{smooth}}\mathcal{L}_{\mathrm{smooth}}
    \nonumber\\
    &+
    \lambda_{\mathrm{event}}\mathcal{L}_{\mathrm{event}}
    +\lambda_{\mathrm{phys}}\mathcal{L}_{\mathrm{phys}}.
    \label{eq:dynamic-objective}
\end{align}
The observation term compares rendered masks, boxes, axes, and functional
landmarks with tracked image evidence. The geometry term anchors visible
object regions to metric point maps, while the smoothness term regularizes
translational and rotational motion over time. The contributions of visual
refinement, depth-derived initialization, and temporal regularization are
evaluated separately in Appendix~\ref{app:ablations}, with trajectory implementation details in Appendix~\ref{app:trajectory-details}.

Procedure structure enters through
\begin{equation}
    \mathcal{L}_{\mathrm{event}}
    =
    \sum_{e_k\in\mathcal{E}}
    c_k\,
    \mathcal{L}_{\tau_k}
    \left(
        \mathcal{Q}_{t_k^s:t_k^e};
        \mathcal{P}_k,r_k
    \right),
    \label{eq:event-loss}
\end{equation}
which imposes event-specific spatial and temporal relations over each
active interval. A relation may require ordered phases such as approach,
entry, dwell, and withdrawal rather than proximity to a target alone.
The physical term separately enforces rigid dimensions, articulation
limits, support, and non-penetration. Together, these terms constrain
trajectories that can agree with image evidence while violating the
modeled interaction geometry. Appendix~\ref{app:event-constraints}
specifies the event constraints, including single-axis hinged motion;
Appendix~\ref{app:pipetting-events} gives the pipetting conditions used
by the corresponding reconstruction diagnostics.

The recovered assets, metric scene, camera state, object trajectories,
joint states, and event list are compiled into an editable 3D
representation (Appendix~\ref{app:world-representation}). The reconstructed procedure can be replayed from new
viewpoints and its geometry, motion, articulation, and event timing can
be inspected or modified independently, providing structured state for
downstream world-model and embodied-agent research.

\section{Evaluation Protocol}
\label{sec:experiments}

We evaluate reconstruction along two complementary dimensions:
image-space agreement with the observed manipulation and, for pipetting,
satisfaction of modeled physical interaction requirements. The latter
measure whether reconstructed trajectories satisfy explicit geometric
constraints, not whether hidden source events are independently verified.

\subsection{Image-Space Reconstruction Evaluation}

\label{sec:evaluation}

\myparagraph{Evaluation setup.}
Camera alignment is estimated from static scene evidence and frozen before
evaluating manipulated-object motion. Reference validity is determined
independently of reconstruction error: observations are excluded only when
the required reference is unavailable or unreliable, never because a
prediction performs poorly. Metrics are computed within each video and
then averaged equally across videos. Additional validity and camera-error
diagnostics are provided in Appendix~\ref{app:evaluation-details}.

\myparagraph{Pipetting.}
We evaluate pipette reconstruction using Mask IoU between the visible SAM3 reference mask $M_t$ and rendered pipette silhouette $\widehat{M}_t$.
The pipette has a thin projected silhouette, making IoU sensitive to small localization and orientation errors.
The reference mask contains only the visible pipette and may be incomplete under occlusion.
Because SAM3 masks also provide observations to \textsc{LabWorld}, this metric measures reconstruction agreement with the source video rather than generalization to held-out annotations.

\myparagraph{Vortexing.}
Hand occlusion makes silhouette IoU unsuitable for much of the vortexing
sequence. We instead use \textbf{Tube-Length--Normalized PCK-AUC}. For
predicted full-tube center $p_i$, source-assisted reference center $q_i$,
and projected reference tube length $L_i$, define
\begin{equation}
    r_i=\frac{\lVert p_i-q_i\rVert_2}{L_i},
    \qquad
    \operatorname{PCK\mbox{-}AUC}
    =
    \frac{1}{N}\sum_{i=1}^{N}\max(0,1-r_i).
    \label{eq:pck-auc}
\end{equation}
The benchmark contains 385 scheduled observations across ten videos,
including partially and fully occluded observations. Reference centers
estimate the midpoint of the complete tube, including its hidden portion,
and projected tube lengths are fixed per video at 30--32 pixels in the
$960\times600$ evaluation image. These source-assisted references are
provisional rather than independently verified ground truth; the metric
therefore evaluates 2D full-tube center localization, not shape,
orientation, 3D pose, or physical interaction.

\myparagraph{Visibility analysis.}
For pipetting, we additionally characterize performance as visible evidence
decreases. Let
\begin{equation}
    r_t =
    \frac{A_t^{\mathrm{SAM3}}}
         {\max(A_t^{\mathrm{render}},1)},
    \qquad
    v_t =
    \operatorname{clip}\!\left(
    \frac{r_t}{\max(Q_{0.75}(r),10^{-6})},0,1
    \right),
    \qquad
    o_t=1-v_t,
    \label{eq:visibility-proxy}
\end{equation}
where $A_t^{\mathrm{SAM3}}$ and $A_t^{\mathrm{render}}$ are the pixel areas
of the reference mask $M_t$ and the rendered silhouette $\widehat{M}_t$,
and $Q_{0.75}(r)$ is the within-video 75th percentile of $r_t$.
We use $v_t\geq0.7$ for the higher-visibility group and also analyze the
continuous association between $o_t$ and IoU. Because this proxy depends
on the predicted silhouette, it is not a calibrated measure of physical
occlusion.

\subsection{Baseline Comparison Protocol}

\label{sec:baseline_protocol}

We compare methods through their final projected reconstructions while
retaining each method's native geometry, camera, scale, and trajectory.
Reconstructions are frozen before scoring, with no post-hoc spatial or
temporal alignment to evaluation references and no score-driven output
selection. No baseline receives \textsc{LabWorld} trajectories, event
constraints, or evaluation reference points. Exact baseline inputs and
implementations are given in Appendix~\ref{app:baseline-details}.

GPT-6 Astra reconstructs each episode independently from the source RGB video.
Do as I Do is evaluated using its monocular object-reconstruction and 6-DoF tracking pipeline, and FoundationPose using an RGB-adapted FoundationPose+SAM3D+MoGe pipeline.

\subsection{Physical Constraint-Satisfaction Diagnostics}

\label{sec:physical_checks}

\begin{figure*}[t]
    \centering
    \includegraphics[width=1\textwidth]{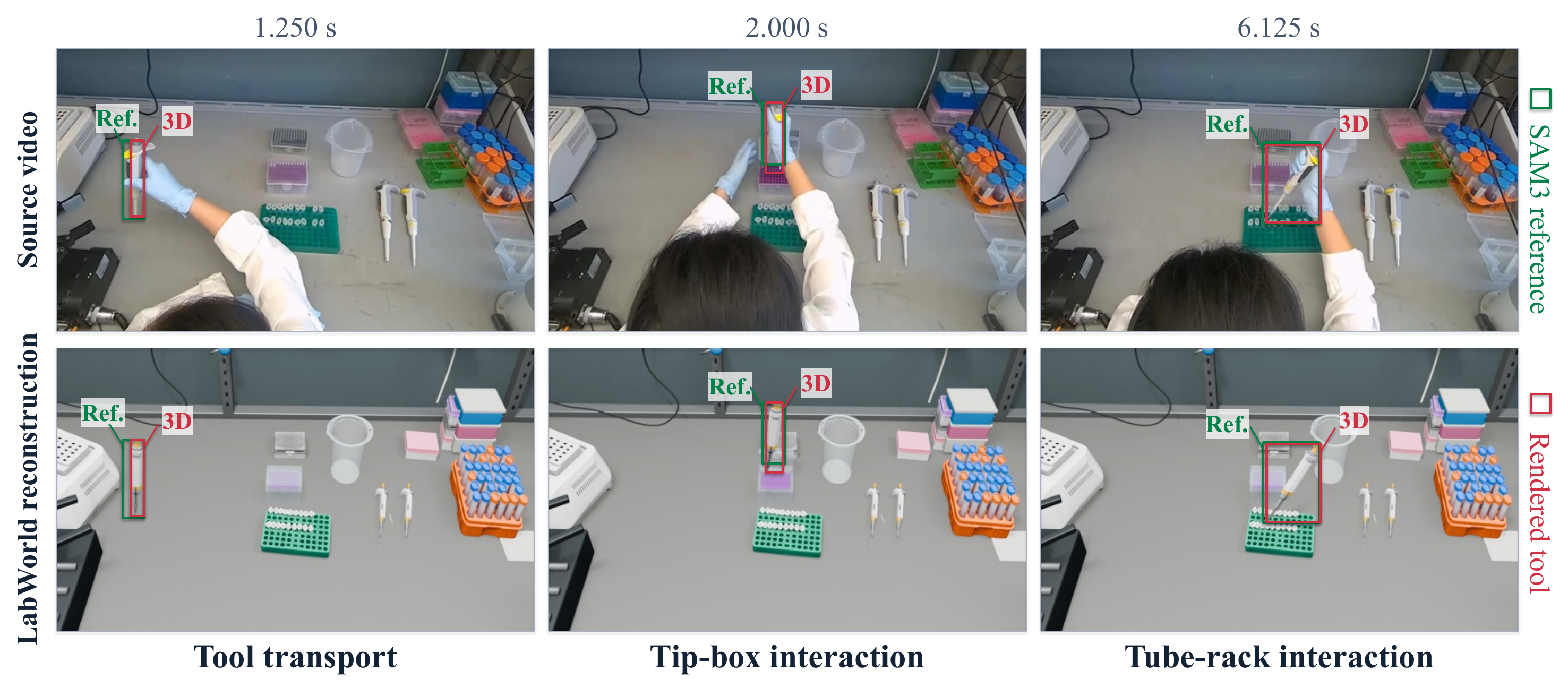}

    \caption{\textbf{Source-video and reconstruction comparison.}
    Source frames (top) and LabWorld reconstructions (bottom) from pipetting
    video 108 are shown at matched times using the same tabletop crop. The
    columns show tool transport, tip-box interaction, and tube-rack interaction.
    Green boxes (Ref.) enclose SAM3 visible reference masks, while red boxes
    (3D) enclose rendered tool silhouettes projected through the fixed
    reconstruction camera. The overlays support comparison of projected
    position and extent, although occlusion can affect the visible reference
    masks. These selected frames illustrate reconstruction states rather than
    verify complete events or liquid transfer.}
    \label{fig:event-box-comparison}

\end{figure*}

Fig.~\ref{fig:event-box-comparison} compares source observations with
projected reconstructions at selected interaction times. Such image-space
comparisons alone do not establish whether the reconstructed motion
satisfies the required 3D interaction constraints.

For the pipetting ablations, we therefore evaluate seven requirements from
the saved 3D scene at 24~Hz: valid cup approach, entry, dwell, withdrawal,
cup collision avoidance, final table support, and global floor clearance.
Event windows, geometry, and tolerances are fixed before the ablation
comparisons. Across five videos, this yields 35 checks per condition.
Exact definitions and numerical thresholds are given in
Appendix~\ref{app:physical-checks}.

These diagnostics deliberately overlap with requirements enforced during
event-conditioned reconstruction. Each evaluated video, however, records
a correctly executed procedure whose true motion satisfies these
requirements, so a reconstruction that violates them contradicts a known
property of the source demonstration. The diagnostics therefore measure
consistency with the demonstrated interaction, not independent recovery of
hidden source events. Passing all checks does not establish liquid
transfer, exact hidden contact, or overall task completion.

\subsection{Ablation Protocol}

\label{sec:ablation_protocol}

We ablate event-conditioned reconstruction, temporal regularization,
depth-derived initialization, and iterative visual refinement on a
prespecified subset comprising the first five pipetting videos
(IDs 101--105), selected before any ablation results were generated.
Each condition uses the same observations, reference masks, camera,
asset geometry, and evaluation procedure, except for the ablated
component, with one reconstruction per condition and video.

At a high level, the event ablation removes event-conditioned interaction
planning while retaining generic physical constraints; the temporal
ablation removes trajectory smoothing and temporal continuity terms; the
depth ablation replaces depth-derived heights with a fixed 0.20\,m
above-table prior; and the visual-refinement ablation retains mask-derived
initialization but removes iterative mask-guided trajectory refinement.
Exact interventions are specified in Appendix~\ref{app:ablation-details}.

All variants are evaluated with the same Mask IoU and physical
constraint-satisfaction protocols. We additionally report trajectory
acceleration as a motion-roughness diagnostic for the temporal ablation.

\section{Results}
\label{sec:results}

We evaluate \textsc{LabWorld} on ten pipetting and ten vortexing videos
against general-purpose agentic and dedicated reconstruction baselines.
The two procedures probe complementary reconstruction regimes: pipetting
evaluates the silhouette of a thin manipulated tool, while vortexing
evaluates localization of a small tube frequently occluded by the
manipulating hand. Evaluation metrics, baseline inputs, and scoring
protocols are defined in Sec.~\ref{sec:experiments}.

\subsection{Reconstruction Accuracy}

\label{sec:reconstruction_results}

\myparagraph{Pipetting.}
\textsc{LabWorld} achieves the highest mean Mask IoU, \textbf{0.501}, compared with 0.468 for Do as I Do, 0.156 for GPT-6 Astra, and 0.026 for FoundationPose (Table~\ref{tab:reconstruction_results}).
Relative to GPT-6 Astra, \textsc{LabWorld} improves mean IoU by
\textbf{0.345} and achieves higher mean IoU on all ten videos.
This evaluation is challenging because the pipette has a thin projected
silhouette, for which small localization or orientation errors produce
substantial changes in mask overlap.

\begin{table}[tb]

    \centering
    \caption{
        \textbf{Reconstruction accuracy across baselines.} Pipetting is evaluated
        using mean Mask IoU and vortexing using Tube-Length--Normalized
        PCK-AUC. Higher is better.
    }
    \vspace{4pt}
    \label{tab:reconstruction_results}
    \resizebox{0.6\textwidth}{!}{
    \begin{tabular}{lcc}
        \toprule
        Method
        & Pipetting IoU $\uparrow$
        & Vortexing PCK-AUC $\uparrow$ \\
        \midrule
        GPT-6 Astra
        & 0.156 & 0.479 \\
        FoundationPose
        & 0.026 & 0.232 \\
        Do as I Do
        & 0.468 & 0.603 \\
        \midrule
        \textsc{LabWorld}
        & \textbf{0.501} & \textbf{0.638} \\
        \bottomrule
    \end{tabular}}

\end{table}

\myparagraph{Vortexing.}
On the ten vortexing videos, \textsc{LabWorld} achieves the highest
Tube-Length--Normalized PCK-AUC of \textbf{0.638}, compared with
0.603 for Do as I Do, 0.479 for GPT-6 Astra, and 0.232 for
FoundationPose.  PCK-AUC evaluates projected 2D center localization
rather than tube shape, orientation, 3D pose, or interaction geometry
and should not be interpreted as a complete measure of 3D reconstruction
quality. This distinction is particularly important because the projected
full-tube lengths are only approximately 30--32 pixels.

Overall, \textsc{LabWorld} achieves the highest mean reconstruction accuracy on both procedures. The two
metrics evaluate different reconstruction properties, and their absolute
values are not directly comparable across tasks.

\subsection{Reconstruction Under Reduced Visibility}

\label{sec:occlusion_results}

\begin{figure}[tb]
    \centering
    \includegraphics[width=\linewidth]{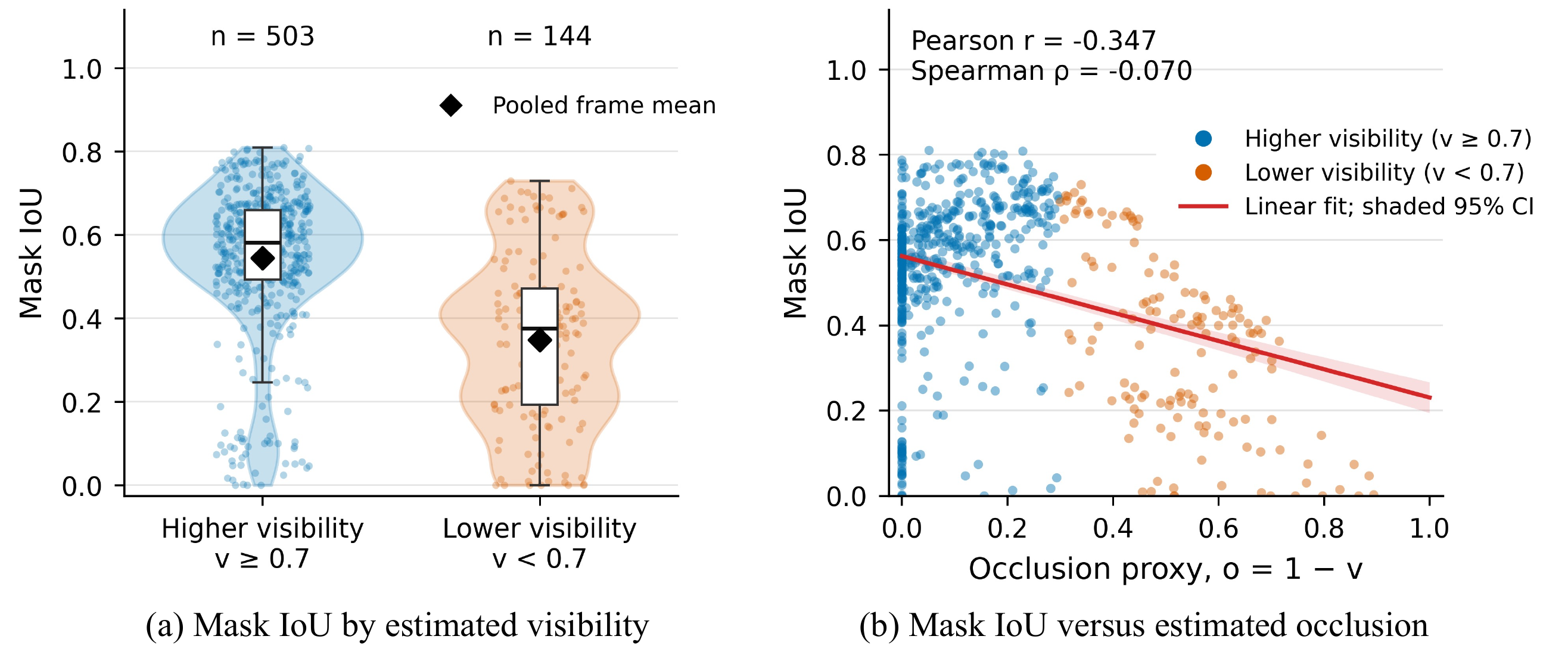}

    \caption{
        \textbf{Pipette reconstruction agreement under reduced visibility.}
        (a) Mask IoU for higher-visibility
        ($v \geq 0.7$, $n=503$) and lower-visibility
        ($v<0.7$, $n=144$) observations across ten
        \textsc{LabWorld} reconstructions. Points denote observations,
        violins show their distributions, boxes indicate medians and
        interquartile ranges, and black diamonds mark pooled means
        (0.545 and 0.349).
        (b) Mask IoU versus the occlusion proxy $o=1-v$ across all
        647 observations. The red line shows the pooled linear fit with
        a pointwise 95\% confidence interval from whole-video resampling.
        Pearson $r=-0.347$ and Spearman $\rho=-0.070$ show that the
        group-level difference does not correspond to a consistently
        monotonic relationship. Visibility is a prediction-dependent
        proxy, not a calibrated measure of physical occlusion.
    }
    \label{fig:iou_distribution}
\end{figure}

We next examine how pipette reconstruction agreement changes as visual
evidence becomes less complete. Using the relative visibility proxy defined
in Sec.~\ref{sec:evaluation}, higher-visibility observations
($v \geq 0.7$) achieve a mean Mask IoU of \textbf{0.545}
($n=503$), compared with \textbf{0.349} for lower-visibility
observations ($v<0.7$; $n=144$), across 647 observations from ten
videos (Fig.~\ref{fig:iou_distribution}a).

The continuous relationship is more nuanced. Mask IoU has a negative
linear association with the occlusion proxy (Pearson $r=-0.347$), but
only a weak rank association (Spearman $\rho=-0.070$)
(Fig.~\ref{fig:iou_distribution}b). Reduced estimated visibility therefore
coincides with lower reconstruction agreement on average but does not
explain reconstruction error by itself. Camera, geometry, and trajectory
errors can also reduce IoU, and both the visibility proxy and IoU depend
on the predicted silhouette.

\subsection{Ablation Study}

We isolate event-conditioned reconstruction, the primary component that
distinguishes \method{} from purely visual trajectory fitting. Table~\ref{tab:event_ablation}
shows that removing event constraints increases Mask IoU from 0.508 to
0.572, but causes physical constraint satisfaction to collapse from
35/35 to 5/35 checks. Thus, higher image-space overlap alone does not
imply a more faithful reconstruction: event conditioning is necessary to
preserve the approach--entry--dwell--withdrawal interaction structure of
the experiment.

\begin{table}[tb]
\centering
\caption{
\textbf{Ablation of event-conditioned reconstruction.}
Removing event constraints improves image-space Mask IoU but causes
physical constraint satisfaction to collapse, demonstrating that visual
agreement alone does not ensure a physically valid reconstruction.
The complete component-wise ablation is reported in
Appendix~\ref{app:ablations}.
}
\vspace{3pt}
\label{tab:event_ablation}
\resizebox{0.8\textwidth}{!}{
\begin{tabular}{lcc}
\toprule
Method & Mask IoU $\uparrow$ & Physical constraints $\uparrow$ \\
\midrule
\method{} & 0.508 & \textbf{35/35} \\
w/o event-conditioned reconstruction & \textbf{0.572} & 5/35 \\
\bottomrule
\end{tabular}}

\end{table}

A complete component-wise ablation, including temporal regularization,
depth-derived initialization, and iterative visual refinement, together
with implementation details and analysis of each component, is provided
in Appendix~\ref{app:ablations}.

\myparagraph{Summary.}
The ablations separate the roles of the reconstruction components:
visual refinement and depth initialization improve image agreement,
temporal regularization improves motion smoothness, and event conditioning
preserves the modeled 3D interaction structure. Together with the baseline
comparisons, these results show that laboratory reconstruction requires
both accurate image-space alignment and an explicit representation of
physically meaningful object interactions.

\section{Conclusion}
With \method{}, we demonstrate that videos of humans performing laboratory experiments can serve as a valuable source for constructing interactive 4D worlds.
\method{} converts recorded laboratory scenes and procedures into an explicit, editable representation, allowing scene geometry, object motion, and procedure events to be inspected and modified.
Event-conditioned reconstruction further incorporates the spatial and temporal requirements of experimental interactions into motion reconstruction, accounting for both image agreement and the interaction structure of the procedure.
Evaluation on pipetting and vortexing videos, together with component ablations, provides empirical support for the framework and the role of event constraints.
By turning human experimental demonstrations into reusable and editable 4D worlds, \method{} offers a way to construct environments for physical intelligence research in laboratories and provides a foundation for future robot learning and experimental planning.

\textbf{Limitation.}
Severe hand occlusion can increase uncertainty in geometry and motion estimation.
The effectiveness of event constraints also depends on the accuracy of event extraction and scene geometry estimation.

\clearpage

\bibliographystyle{plainnat}
\bibliography{lab2world_references}

\clearpage

\appendix

\section{Agent Workflow and Reproducibility Scope}
\label{app:automation-scope}

The fully automatic workflow includes agent preparation of the reconstruction
inputs and event list, followed by asset generation, scene reconstruction,
motion reconstruction, and verification. GPT-6 supplies the input configuration
and event definitions from video observations. The term refers to automation
of this complete workflow rather than execution of a solver from an externally
prepared configuration.

The current 105--108 reproduction package starts from cached SAM3 and
MapAnything observations, reusable assets, and saved configurations containing
camera, layout, scale, articulation, and event information. These runs test
reproducibility from the declared inputs. They do not independently establish
fully automatic reconstruction from raw video. A reviewed configuration alone
does not identify whether its preparation or correction was performed by an
agent or a person. End-to-end evaluation must record input provenance, agent
actions, and any human intervention separately. Inspection of a completed
output must also be distinguished from changes to reconstruction inputs.

\section{Executable World Representation}
\label{app:world-representation}

The reconstructed world is stored as explicit, independently inspectable
layers rather than a monolithic rendered video.  For an episode with $T$
frames, the exported state contains:
\begin{equation}
    \mathcal{W}
    =
    \left(
    \mathcal{A},\mathcal{S},
    \{K_t,T^W_{C,t}\}_{t=1}^{T},
    \{T^W_{i,t}\}_{i,t},
    \{q_{j,t}\}_{j,t},
    \mathcal{E}
    \right).
\end{equation}
Here, $\mathcal{A}$ is the reusable asset library, $\mathcal{S}$ is the static
metric laboratory scene, $K_t$ and $T^W_{C,t}$ are camera intrinsics and
extrinsics, $T^W_{i,t}\in\SEthree$ is the pose of rigid object $i$, $q_{j,t}$
is the state of articulated joint $j$, and $\mathcal{E}$ is the scientific
event list.  This representation allows geometry, motion, articulation, and
event timing to be audited separately.  It also permits a downstream agent to
change an initial object state or action while retaining the remainder of the
reconstructed laboratory.

\section{Event Vocabulary and Geometric Constraints}
\label{app:event-constraints}

Table~\ref{tab:event-constraints} summarizes how the agent-generated event
list supplies geometric conditions for motion reconstruction.
The event vocabulary is intentionally defined in terms of functional
relations, so the same constraint can be instantiated for previously unseen
assets with compatible functional parts.

\begin{table}[h]
    \centering
    \small
    \setlength{\tabcolsep}{5pt}
    \caption{Scientific events and the geometric relations they instantiate.}
    \begin{tabular}{>{\raggedright\arraybackslash}p{1.9cm}>{\raggedright\arraybackslash}p{3cm}>{\raggedright\arraybackslash}p{9.4cm}}
        \toprule
        Event & Functional parts & Reconstruction constraint \\
        \midrule
        Grasp & handle, hand & Object and hand motion become locally coupled,
        and the object is released from its previous support. \\
        Transport & rigid object & Preserve object dimensions and orientation
        continuity while allowing free motion above the table. \\
        Dispense & tool tip, opening & Enforce the required approach, entry,
        dwell, and withdrawal phases while retaining the observed tool axis. \\
        Insert & tube axis, slot & Align axes, constrain lateral offset, and
        enforce a valid insertion depth without penetration. \\
        Open/close & lid, hinge, base & Keep the base supported and restrict the
        lid to a revolute joint with fixed axis and joint limits. \\
        Place & object support, surface & Reach support contact with low terminal
        velocity and remain non-penetrating after release. \\
        \bottomrule
    \end{tabular}
    \label{tab:event-constraints}
\end{table}

Event boundaries need not be treated as hard discontinuities.  For an event
active over $[t_s,t_e]$, we use a temporally tapered confidence $w_t$ that rises
before the stable interaction phase and decays after release.  A dispensing
proximity term, for example, can be written as
\begin{equation}
    \mathcal{L}_{\mathrm{dispense}}
    =
    \sum_{t=t_s}^{t_e} w_t
    \left\|
        p_{\mathrm{tip},t}-p_{\mathrm{opening}}
    \right\|_2^2,
\end{equation}
with an additional orientation term when the receptacle defines a preferred
approach axis. This proximity term is not a complete event constraint or a
completion test. It must be combined with the phase-specific depth, timing,
and non-penetration conditions below. Tapering reduces abrupt changes at an
event boundary but does not by itself guarantee trajectory continuity.

For an open or close event on an articulated container, the base remains
supported while the lid follows a single-axis revolute joint
\begin{equation}
    T^W_{\mathrm{lid},t}
    =
    T^W_{\mathrm{base}}
    T_{\mathrm{hinge}}(q_t),
    \qquad q_{\min}\le q_t\le q_{\max}.
    \label{eq:hinge}
\end{equation}

\subsection{Pipetting Event Definitions}

\label{app:pipetting-events}

For the evaluation protocol, each video has at most ten task events. A
complete interaction is counted once even when it contains several phases.
For example, beaker entry, dwell, and withdrawal form one event rather than
three independent successes. Sequence-wide collision checks and visual
diagnostics remain separate from the task event count. Existing engineering
checklists must be regrouped under these rules before reporting benchmark
event completion. Each entry records source timestamps, frame indices,
participants, required phases, and the tolerances used for evaluation.

\myparagraph{Beaker interaction.}
The pipette tip must enter through the opening, remain inside for the required
dwell interval, and withdraw through the opening before moving away. Entry
requires crossing the opening plane within its interior and reaching below
the rim without intersecting the vessel wall. The evaluator checks phase
order and timing as well as geometry. The current pipetting planner uses a
24 fps trajectory and a configured shallow insertion depth of 8 mm. This
depth is an explicit reconstruction constraint, not a measurement of the
hidden tip in the source video. Dwell duration and timing tolerances must be
recorded for each event. The event evaluates pipette motion, not liquid
transfer.

\myparagraph{Rack and tip-box interactions.}
The green target in the pilot videos is a rack with raised white tube
openings, not a solid plate. Reaching the rack region does not establish
contact with a particular opening. Likewise, entering an open tip-box region
does not establish insertion into a specific slot. The checklist distinguishes
visible approach from exact contact or insertion. An occluded contact remains
unknown when the reference evidence cannot establish the required relation.

\myparagraph{Final placement and collision checks.}
Final placement requires the complete tool mesh to reach table support and
remain stable through the final rendered frame. The reference checklist must
not require visible hand release when the source video does not show it.
The saved Blender scene is reopened to inspect the applied trajectory rather
than only the planned positions. Tool--beaker intersection and table
penetration are checked separately over the rendered sequence. These checks
do not certify continuous collision freedom for all objects in the scene.

\section{Trajectory Reconstruction Details}
\label{app:trajectory-details}

For a rigid, elongated laboratory tool, a visible-mask centroid is not a
stable pose observation because hand occlusion changes which portion of the
object is visible.  We instead maintain semantic endpoints or functional landmarks when
available.  Given two image rays $r_{1,t}$ and $r_{2,t}$ and a canonical object
length $\ell$, the lifted endpoints are constrained by
\begin{equation}
    p_{1,t}=o_t+d_{1,t}r_{1,t},\qquad
    p_{2,t}=o_t+d_{2,t}r_{2,t},\qquad
    \|p_{2,t}-p_{1,t}\|_2=\ell.
\end{equation}
MapAnything point maps initialize the depths $d_{1,t}$ and $d_{2,t}$, and the
fixed length rejects depth combinations that would deform the tool.  A third
non-collinear landmark, an asymmetric hook, or a canonical mesh frame is
required to determine roll around the long axis.  When this evidence is
occluded, roll is propagated temporally rather than re-estimated from an
unstable silhouette.

The final sequence optimizer operates on correction variables relative to the
geometric initialization.  Translation and rotation corrections are coupled
across time using first- and second-order penalties.  Event and physical terms
are then applied after image-space alignment, so increasing silhouette overlap
cannot move a tool away from a high-confidence receptacle contact or through a
support surface.

\section{Evaluation and Validity Masks}
\label{app:evaluation-details}

The reference moving-object mask is obtained independently from the rendered
object-ID mask.  Evaluation uses a single camera alignment estimated from
static scene evidence and frozen for the complete sequence.

A per-frame validity flag records whether the manipulated object is observable
and whether reference annotation is trustworthy.  It is determined without
access to the reconstruction error.  Frames are not removed because IoU is
zero or because a predicted trajectory is visually poor.  We report both the
metric over valid frames and the validity coverage
\begin{equation}
    \operatorname{Coverage}
    = \frac{|\mathcal{T}_{\mathrm{valid}}|}{T}.
\end{equation}

The oracle global translation is
defined as
\begin{equation}
    \Delta^{*}
    =
    \arg\max_{\Delta}
    \frac{1}{|\mathcal{T}_{\mathrm{valid}}|}
    \sum_{t\in\mathcal{T}_{\mathrm{valid}}}
    \operatorname{IoU}
    \left(M_t,
    \operatorname{Translate}(\widehat M_t,\Delta)\right).
    \label{eq:oracle-shift}
\end{equation}
The same diagnostic translation is used for mask and box metrics.  It is
retained only as a diagnostic upper bound for camera error and is never used
to modify the reconstructed trajectory.

\section{Current Limitations}
\label{app:limitations}

The current system has several important limitations.  Generated assets can
miss fine functional geometry, transparency, material properties, or exact
physical dimensions.  Monocular metric reconstruction remains uncertain for
thin and reflective tools, particularly during severe hand occlusion.  Event
detection reduces ambiguity but an incorrect event can impose an incorrect
geometric relation.  Contact and non-penetration constraints do not yet replace
full system identification of mass, friction, compliance, and fluid dynamics.
Finally, the present benchmark focuses on laboratory tabletop procedures. The
extent to which its object and event vocabulary transfers to substantially
different scientific domains remains an open question.

\section{Baseline Evaluation Details}
\label{app:baseline-details}

All baselines produce one frozen reconstruction per episode rather than
selecting the best output according to evaluation score. Comparisons use
matched source timestamps under the corresponding task protocol. Metrics
are computed within each video and then averaged equally across videos;
missing or incompatible outputs are not silently removed or replaced.

\myparagraph{Agentic reconstruction.}
GPT-6 Astra reconstructs each episode independently from its full source RGB video without access to \textsc{LabWorld} outputs, SAM3 evaluation masks, MapAnything geometry, or scoring reference points.

\myparagraph{Do as I Do.}
For pipetting, Do as I Do receives the complete source RGB video, one
midpoint reference frame with two positive pipette prompts, and the text
prompt ``right hand.'' Its reconstruction pipeline recovers an object
mesh, camera, scale, and per-frame 6-DoF poses. The recovered mesh is
rendered using the method's own camera and poses, without access to a
\textsc{LabWorld} asset, camera, MapAnything depth, or evaluation SAM3
mask. The same frozen-reconstruction principle is used for vortexing.

\myparagraph{FoundationPose.}
We evaluate an RGB-adapted FoundationPose+SAM3D+MoGe pipeline. Source
video is sampled at 8~Hz using nearest native frames and resized to
$960\times600$. FoundationPose receives independently predicted MoGe
depth and camera intrinsics, a fresh SAM3D mesh with its inferred scale,
and a SAM2 mask for initial registration. SAM3D is initialized from a
selected visible source frame and its SAM2 segmentation, so offline
initialization may use a later visible frame. It receives no
\textsc{LabWorld} asset, camera, trajectory, event constraints,
MapAnything depth, or evaluation SAM3 mask.

We use the official FoundationPose registration and tracking procedure
with five registration iterations and two tracking iterations. The
resulting object-to-camera transforms and recovered SAM3D mesh are rendered
using the predicted poses and MoGe intrinsics. One frozen trajectory is
reported per video. Source-based initialization corrections were permitted
before scoring, but no score-driven pose corrections were made. Pipetting
and vortexing outputs are scored under the corresponding shared protocols.

\section{Physical Constraint-Satisfaction Definitions}
\label{app:physical-checks}

For the pipetting ablations, seven physical requirements are recomputed
directly from the final saved 3D scene at 24~Hz rather than taken from
internal planner success flags:

\begin{enumerate}
    \item \textbf{Cup approach:} the final approach sample places the tip
    at least 3~mm above the rim and within the cup radius minus 6~mm.

    \item \textbf{Entry:} the tip crosses the rim plane downward through
    the inset opening during the prescribed entry window.

    \item \textbf{Dwell:} the tip remains within the inset opening, more
    than 2~mm below the rim and more than 4~mm above the configured cup
    base, for at least two consecutive samples spanning $1/24$~s.

    \item \textbf{Withdrawal:} the tip crosses upward through the inset
    opening without subsequent re-entry and finishes at least 3~mm above
    the rim or 20~mm radially beyond the cup radius.

    \item \textbf{Cup collision avoidance:} no pipette--cup triangle
    intersection occurs at any output frame and no consecutive tip segment
    intersects the cup surface.

    \item \textbf{Final table support:} the lowest pipette vertex remains
    within 1~mm of the configured 0.5~mm table clearance throughout the
    terminal-support window.

    \item \textbf{Global floor clearance:} all pipette vertices remain
    above the table plane up to a $10^{-6}$~m numerical tolerance.
\end{enumerate}

Event windows, scene geometry, and numerical tolerances are fixed before
the ablation comparisons and remain identical across conditions. Seven
requirements over five videos yield 35 checks per condition.

These diagnostics overlap with contact, support, and collision requirements
enforced during event-conditioned reconstruction. The evaluation videos,
however, are collected in our laboratory and record correctly executed
procedures. In every evaluated demonstration the operator's actual motion
enters the vessel through its opening, dwells, withdraws, avoids collision
with the vessel, and returns the tool to stable table support, so the seven
requirements are known to hold in the true motion of each video. A
reconstruction that violates them therefore contradicts a known property of
the demonstrated procedure, while satisfying them indicates consistency
with the demonstration rather than exact metric recovery of it. The
diagnostics thus quantify agreement with known properties of the source
motion rather than independently verified source-event correctness.

\section{Ablation Implementation Details}
\label{app:ablation-details}

The ablation study uses the first five pipetting videos (IDs 101--105),
a prespecified subset selected before any ablation results were generated.
All ten videos had completed full-pipeline evaluations before the subset
was specified. Unless stated otherwise, each condition preserves the source
observations, reference masks, camera calibration, asset geometry,
source-frame alignment, and evaluation procedure, and removes only the
specified component. All conditions use identical optimization weights,
constraint tolerances, and event windows across videos, with no
per-video tuning.

\myparagraph{Event-conditioned reconstruction.}
The event ablation bypasses event-conditioned contact-region and
orientation projections, prescribed terminal rest, and dense
cup-interaction planning, including cup-avoidance corrections. Generic
floor nonpenetration and the reviewed lid animation remain active. This
constitutes a functional removal of event-conditioned reconstruction rather
than simply setting an event-loss weight to zero.

\myparagraph{Temporal regularization.}
The temporal ablation removes seed-trajectory smoothing, the temporal
optimization penalty, neighboring-pose motion repairs, and continuity
restrictions during candidate selection. Prepared observations, per-frame
initialization priors, interpolation, and event planning remain unchanged.

\myparagraph{Depth-derived initialization.}
The depth ablation replaces MapAnything-derived per-frame heights and
their downstream use in motion limits with a fixed 0.20\,m above-table
height for every observation and video. This fallback prior is fixed
before execution and is not selected through evaluation-score tuning.

\myparagraph{Iterative visual refinement.}
The visual-refinement ablation retains SAM masks and mask-derived
initialization, including anchors, axes, extents, depth-derived heights,
initialization smoothing, and event constraints, but bypasses iterative
mask/anchor optimization, endpoint refinement, and mask-guided candidate
selection. It therefore removes iterative visual refinement rather than
all visual information.

\myparagraph{Motion-roughness diagnostic.}
For the temporal ablation, motion smoothness is characterized by the mean
norm of the second finite difference of saved-scene pipette-center
positions, scaled by the squared 24~Hz sampling rate and averaged equally
across videos. This acceleration measure is a trajectory-roughness
diagnostic rather than independent source-motion ground truth.

\section{Extended Ablation Study}
\label{app:ablations}

We report the complete component-wise ablation of \method{} on the five
pipetting episodes described in Appendix~\ref{app:ablation-details}.
Event-conditioned reconstruction is the primary ablation discussed in the
main text; here, we additionally evaluate temporal regularization,
depth-derived initialization, and iterative visual refinement. Full
implementation details for each ablation are provided in
Appendix~\ref{app:ablation-details}.

\begin{table}[h]
\centering
\caption{
\textbf{Complete component-wise ablation of \method{}.}
Mask IoU measures image-space agreement, while physical constraints measure
satisfaction of the modeled experimental interaction geometry. Event
conditioning is required for physical validity despite higher IoU without
it, while the remaining ablations characterize complementary contributions
to visual accuracy and trajectory coherence.
}
\label{tab:full_ablation}
\begin{tabular}{lcc}
\toprule
Method & Mask IoU $\uparrow$ & Physical constraints $\uparrow$ \\
\midrule
\method{} & 0.508 & \textbf{35/35} \\
w/o event-conditioned reconstruction & \textbf{0.572} & 5/35 \\
w/o temporal regularization & 0.514 & \textbf{35/35} \\
w/o depth-derived initialization & 0.487 & \textbf{35/35} \\
w/o iterative visual refinement & 0.223 & \textbf{35/35} \\
\bottomrule
\end{tabular}
\end{table}

\myparagraph{Event-conditioned reconstruction.}
Removing event conditioning increases Mask IoU from 0.508 to 0.572,
but physical constraint satisfaction collapses from 35/35 to 5/35.
The higher image-space agreement therefore does not correspond to a more
faithful reconstruction: without event conditioning, the trajectory fails
to preserve the experimental interaction structure. In contrast, the full
\method{} pipeline preserves the approach--entry--dwell--withdrawal
structure while maintaining strong visual agreement. This demonstrates
that image-space fitting alone is insufficient to recover physically
meaningful experimental trajectories.

\myparagraph{Temporal regularization.}
Removing temporal regularization slightly increases Mask IoU from 0.508
to 0.514 and preserves all 35/35 physical constraints, but increases mean
trajectory acceleration from 5.410 to 6.070\,m/s$^2$ (+12.2\%).
The small improvement in frame-wise alignment, therefore, comes with rougher
motion. Temporal regularization favors a coherent reconstruction of the
procedure over independently maximizing agreement at individual
observations.

\myparagraph{Depth-derived initialization.}
Without depth-derived initialization, Mask IoU decreases from 0.508 to
0.487 while physical constraint satisfaction remains 35/35. Depth
therefore provides a useful spatial initialization that improves subsequent
image-space alignment, while event constraints preserve the required
interaction geometry.

\myparagraph{Iterative visual refinement.}
Removing iterative visual refinement produces the largest decrease in
Mask IoU, from 0.508 to 0.223, while preserving all 35/35 physical
constraints. Thus, geometric and event constraints can maintain the
required interaction structure, but do not by themselves provide precise
image-space localization. Visual refinement complements these constraints
by aligning the physically valid trajectory with the observed object
appearance.

\end{document}